\documentclass{article}
\usepackage{iclr2027_conference,times} \iclrfinalcopy 
\usepackage[T1]{fontenc}
\usepackage[utf8]{inputenc}
\usepackage{amsmath,amssymb,booktabs,graphicx,tabularx}
\usepackage{tikz}
\usetikzlibrary{arrows.meta,positioning}
\usepackage[hidelinks]{hyperref}
\usepackage{url}
\usepackage{wrapfig}
\usepackage{caption}
\title{See, Foresee, Act: Intent-Aware World\\Modeling for Proactive Retail Assistance}

\author{
    \textbf{Honghui Zhang},
    \textbf{Anna Min},
    \textbf{Chenmeinian Guo}, 
    \textbf{Yujia Zhang}, \\ 
    \textbf{Zezhou Zhang},  
    \textbf{Yichen Yu}, 
    \textbf{Guanyu Liu},
    \textbf{Yongming Qin}, 
    \textbf{Chongguo Song}, \\
    \textbf{Mengyue Yang}, 
    \textbf{Lei Yu},
    \textbf{Tianyu Shi}
    \vspace{0.5em} \\
    \textsuperscript{1}Meetvista Research Team\\ \quad
    \vspace{0.5em} \\
    {\small \textbf{Correspondence:} \href{mailto:tianyu.shi3@mcgill.ca}{tianyu.shi3@mcgill.ca}}
}

\newcommand{\hold}{\textsc{Hold}}

\usepackage{longtable,array,calc}

\begin{document}
\maketitle

\begin{abstract}
World models allow agents to anticipate action consequences, but proactive assistance also requires interpreting incomplete observations and deciding whether to intervene before a user asks for help. We propose See--Foresee--Act and instantiate it as the Proactive Intent World Model \textbf{(PIWM) }for retail assistance. See preserves role-relevant observable evidence alongside AIDA--BDI customer states; Foresee predicts one-step consequences for candidate actions; Act explicitly compares those consequences, including non-intervention. GuidanceSalesBench supplies predominantly synthetic supervision linking observations, states, candidate outcomes, and decisions. On 509 Full test examples, consequence-based selection achieves 63.65\% decision accuracy, compared with 45.38\% for direct decision output, an improvement of 18.27 percentage points. These results support candidate-consequence comparison for decision reasoning in this retail task. Code and data will be released after consent verification and upon publication.
\end{abstract}

\section{Introduction}\label{introduction}

\begin{wrapfigure}[13]{r}{0.6\textwidth}
  \vspace{-8pt}
  \centering
  \includegraphics[width=\linewidth]{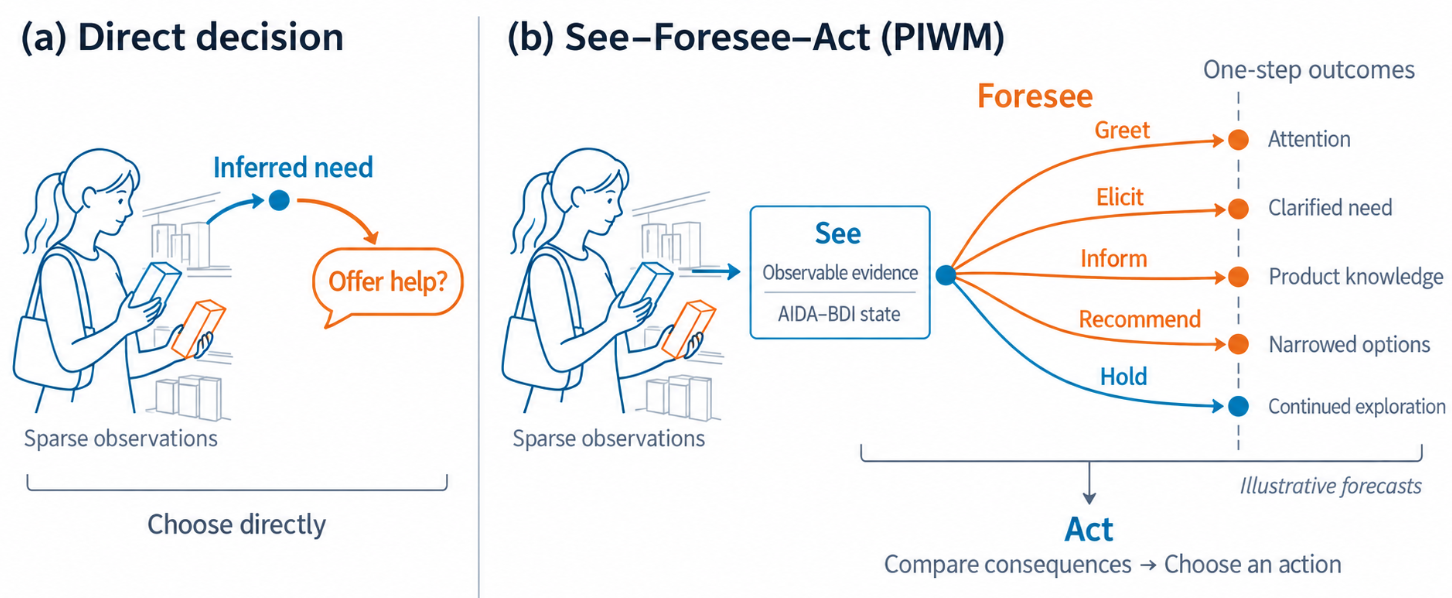}
  \captionsetup{font=footnotesize,labelfont=bf,skip=3pt}
  \caption{Direct decision versus PIWM. See preserves observable evidence and inferred states; Foresee predicts one-step outcomes; Act compares alternatives, including non-intervention (Hold).}
  \label{fig:teaser}
  \vspace{-4pt}
\end{wrapfigure}

World models enable agents to predict action-conditioned state changes
and use these predictions to guide decisions
\citep{hafner2020dreamer, hafner2023mastering}. Proactive assistance
additionally requires inferring service needs from incomplete
observations and deciding whether to intervene before a user makes an
explicit request \citep{yang2026contextagent, horvitz1999principles}. A
retail customer may compare products, inspect labels, or repeatedly
examine the same item. These behaviors may indicate a need for
assistance or independent exploration. We study proactive retail assistance as a decision problem: given sparse pre-interaction observations, an agent must infer the customer’s service needs and decide whether and how to intervene by considering the consequences of both assistance and silence\citep{fu2026prism}.


\begin{figure*}[t]
\centering
\includegraphics[width=\textwidth]{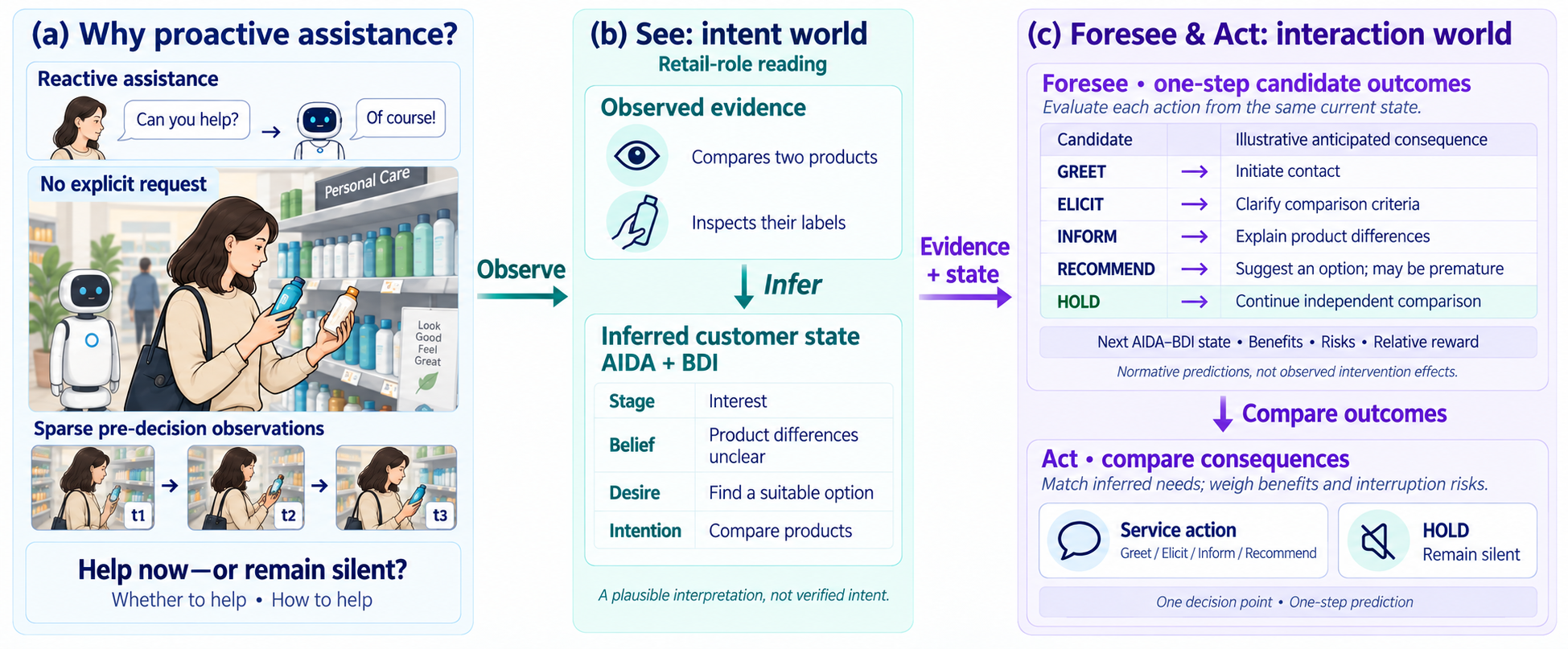}
\caption{Motivation and task scope of PIWM. Proactive retail assistance requires deciding whether and how to help before an explicit customer request. The intent world represents observed human--object evidence and inferred AIDA--BDI states, while the interaction world associates candidate service actions with normative next-state targets. We evaluate sparse-video \emph{See}, action-conditioned \emph{Foresee}, and intervention selection including \hold{} at a single decision point.}
\vspace{-20pt}
\label{fig:piwm-motivation}
\end{figure*}

This problem presents two challenges: grounding customer-state inference in observable evidence and selecting interventions under uncertain service needs. First, observations must be translated into task-relevant
states rather than generic descriptions \citep{yang2026contextagent}.
Intent alone can compress away the evidence needed to distinguish a
request for information from independent comparison: similar inferred
intentions can arise from different gaze, engagement, and
product-handling patterns. Second, action selection must weigh
assistance benefits against uncertain needs and interruption costs
\citep{horvitz1999principles, fu2026prism}. This requires considering
what happens if the agent remains silent as well as what happens if it
acts. We formulate the task at a single decision point, linking sparse-video observations to structured customer states, action-conditioned normative outcomes, and intervention choices that include remaining silent.

We propose the \textbf{See--Foresee--Act framework} and instantiate it
as \textbf{PIWM} (Proactive Intent World Model) for retail assistance. See extracts role-relevant
human--object evidence and predicts a structured customer state from
sparse video. Foresee predicts one-step normative outcomes for each
candidate service action. Act receives the current evidence, inferred
state, and candidate forecasts, then explicitly compares alternatives
including non-intervention (Hold). The common state and action interfaces support both end-to-end assistance decisions and controlled comparisons of evidence use and consequence prediction.

PIWM combines \textbf{AIDA(Attention-Interest-Desire-Action)} purchasing stages \citep{barry1987development} with
\textbf{BDI} fields for\textbf{ beliefs, desires, and intentions} \citep{rao1995bdi},
while retaining explicit evidence about engagement, gaze, posture, and
object interaction. The inferred fields express a task interpretation;
the evidence fields preserve the observations supporting it
\citep{rabinowitz2018machine, bai2025qwen25vltechnicalreport}. Both
enter consequence prediction and action selection. This design lets
downstream modules compare service alternatives against the same task
interpretation and its visible basis.

Hold is modeled through the same consequence interface as Greet, Elicit,
Inform, and Recommend. An available silence option establishes what the
agent may choose; a forecast describes why waiting could preserve
independent exploration or avoid premature interruption. We therefore
test the contribution of the Hold forecast while retaining the action
itself and the non-intervention instruction. This connects the value of
non-intervention to the comparison used for other service actions
\citep{horvitz1999principles, lu2024proactive}.

To support systematic evaluation of this task, we construct \textbf{GuidanceSalesBench}, linking sparse pre-interaction videos, structured customer states, candidate-action outcomes, and normative action labels. Rule-guided, expert-reviewed synthetic examples
supervise observation-to-state, state--action-to-consequence, and
state-to-action mappings. The learned components are composed at
inference by supplying generated candidate consequences to Act.
The benchmark supports component-level evaluation of state inference, consequence prediction, and action selection, together with full-pipeline evaluation (Full) of video-to-action decisions.

On 509 Full test examples, consequence comparison achieves 63.65\% accuracy (324/509), versus 45.38\% (231/509) for direct decision output. This improves accuracy by 18.27 percentage points, a net increase of 93 correct decisions. Removing explicit evidence fields yields 27.70\% (141/509); withholding the Hold forecast yields 47.94\% (244/509). PIWM achieves 62.20\%, 74.43\%, and 89.80\% on See, Foresee, and Act, respectively.



Our main contributions are as follows:

\begin{enumerate}
\item
  \textbf{A task formulation for proactive retail assistance.}
  We study whether and how to intervene from sparse pre-interaction
  observations, connecting observable evidence and inferred customer
  states with action-conditioned normative outcomes at a single
  decision point.

\item
  \textbf{GuidanceSalesBench for structured evaluation.}
  We construct a benchmark linking sparse observations, customer
  states, normative candidate outcomes, and preferred actions,
  supporting both component-level and full-pipeline evaluation.

\item
  \textbf{A See--Foresee--Act framework instantiated as PIWM.}
  The framework connects evidence-grounded state inference,
  one-step candidate-consequence prediction, and comparative
  action selection through common state and action interfaces.

\item
  \textbf{Empirical findings on evidence, consequence comparison,
  and non-intervention.}
  Controlled comparisons show how candidate forecasts, explicit
  evidence fields, and the Hold forecast contribute to decision
  accuracy on GuidanceSalesBench.
\end{enumerate}

\section{Related Work}\label{related-work}

\subsection{Proactive assistance and intervention
timing.}\label{proactive-assistance-and-intervention-timing.}

Mixed-initiative interaction balances assistance benefits against
uncertainty and interruption costs \citep{horvitz1999principles}.
Proactive Agent studies human acceptance and rejection, LlamaPIE
separates response timing from generation, and ProAssist evaluates
streaming visual assistance
\citep{lu2024proactive, chen2025llamapie, zhang2025proassist}.
ContextAgent connects sensory and personal context to service initiation
and tool use \citep{yang2026contextagent}; PRISM studies risk-sensitive,
uncertainty-aware proactive deliberation \citep{fu2026prism}. These
works motivate deciding whether assistance is warranted. PIWM focuses on
a complementary question: how explicit retail evidence and
candidate-specific forecasts inform a shared choice among five
interventions, including Hold.

ReAct and Reflexion combine reasoning, action, and feedback, while
trajectory and process refinement improve behavior through exploration
and supervision
\citep{yao2023react, shinn2023reflexion, song-etal-2024-trial, xiong-etal-2024-watch, wang-etal-2025-steca}.
PIWM evaluates a single decision point rather than an executed
interaction trajectory. Its non-intervention control holds the candidate
set and instruction fixed while removing only the Hold forecast,
directly examining the value of anticipating silence in action
comparison.

\subsection{Intent modeling and structured
state.}\label{intent-modeling-and-structured-state.}

Partial-observation control distinguishes observable inputs from
underlying state \citep{aastrom1965optimal}. Machine theory of mind
models other agents from observed behavior
\citep{rabinowitz2018machine}. AIDA describes purchasing stages and BDI
organizes beliefs, desires, and intentions
\citep{barry1987development, rao1995bdi}. VAGEN separates visual state
estimation and transition modeling within multi-turn world-model
reasoning \citep{wang2025vagen}. PIWM distinguishes two parts of the
current representation: an inferred task state and explicit visible
evidence. Its evidence-removal experiment tests whether the latter
contributes beyond retaining stage, intent, and BDI, rather than
treating state structure alone as sufficient for decisions.

Interactive environments support evaluating actions through execution in
household, scientific, and shopping tasks
\citep{puig2018virtualhome, shridhar2020alfred, Wang2022ScienceWorld, yao2022webshop}.
PIWM instead links state interpretations to candidate interventions at a
single observation point. Its evaluation primarily measures agreement
with a constructed schema, rather than behavioral validation of inferred
mental states or successful interactive task completion.

\subsection{World models and decision
utility.}\label{world-models-and-decision-utility.}

World models support prediction and control through learned transitions
\citep{ha2018world, hafner2020dreamer, hafner2023mastering}. RAP couples
a language-model world model with search, while other language-agent
methods learn or construct planning models from experience and world
knowledge
\citep{hao-etal-2023-reasoning, xiang2024language, guan2023leveraging, qiao2024agent, wang-etal-2025-world, fang-etal-2025-webevolver}.
PIWM learns normative one-step retail outcomes and uses them for
explicit candidate comparison. Its question concerns the information
needed for a proactive service decision, including the consequence of
not intervening.

Closed-loop evaluation links world-model predictions to downstream task
performance \citep{zhang2025world}. We compare approaches along four
dimensions: the information represented from observation, the
supervision for candidate outcomes, the treatment of non-intervention,
and the decision setting. PIWM combines explicit retail evidence with
AIDA--BDI, learns candidate outcomes from synthetic normative
supervision, predicts Hold through the same interface as active actions,
and evaluates single-step decision accuracy. Its direct-decision and
forecast-based conditions test the utility of this combination on the
same Full set.

\section{Method}\label{method}

\subsection{Task definition and
framework}\label{task-definition-and-framework}

PIWM addresses service decisions before an explicit customer request.
Given sparse pre-decision video observations \(o\) and a retail role
description, the agent selects an action from candidates \(C\): Greet,
Elicit, Inform, Recommend, or Hold. The inference pipeline first
constructs a customer state, predicts a one-step consequence for each
candidate, and then explicitly compares those consequences to select an
action.

The three interfaces are:

\[
(\hat z,\hat s)=\mathrm{See}_{\theta}(o),\quad \hat y^{(a)}=\mathrm{Foresee}_{\theta}(\hat z,\hat s,a),\quad \hat a=\mathrm{Act}_{\theta}\!\left(\hat z,\hat s,\{(a,\hat y^{(a)})\}_{a\in C}\right).
\]

Here \(\hat z\) summarizes observable evidence, \(\hat s\) represents
the customer state, and \(\hat y^{(a)}\) describes the anticipated
consequence of candidate \(a\).

\begin{figure}[t]
\centering
\includegraphics[width=\textwidth]{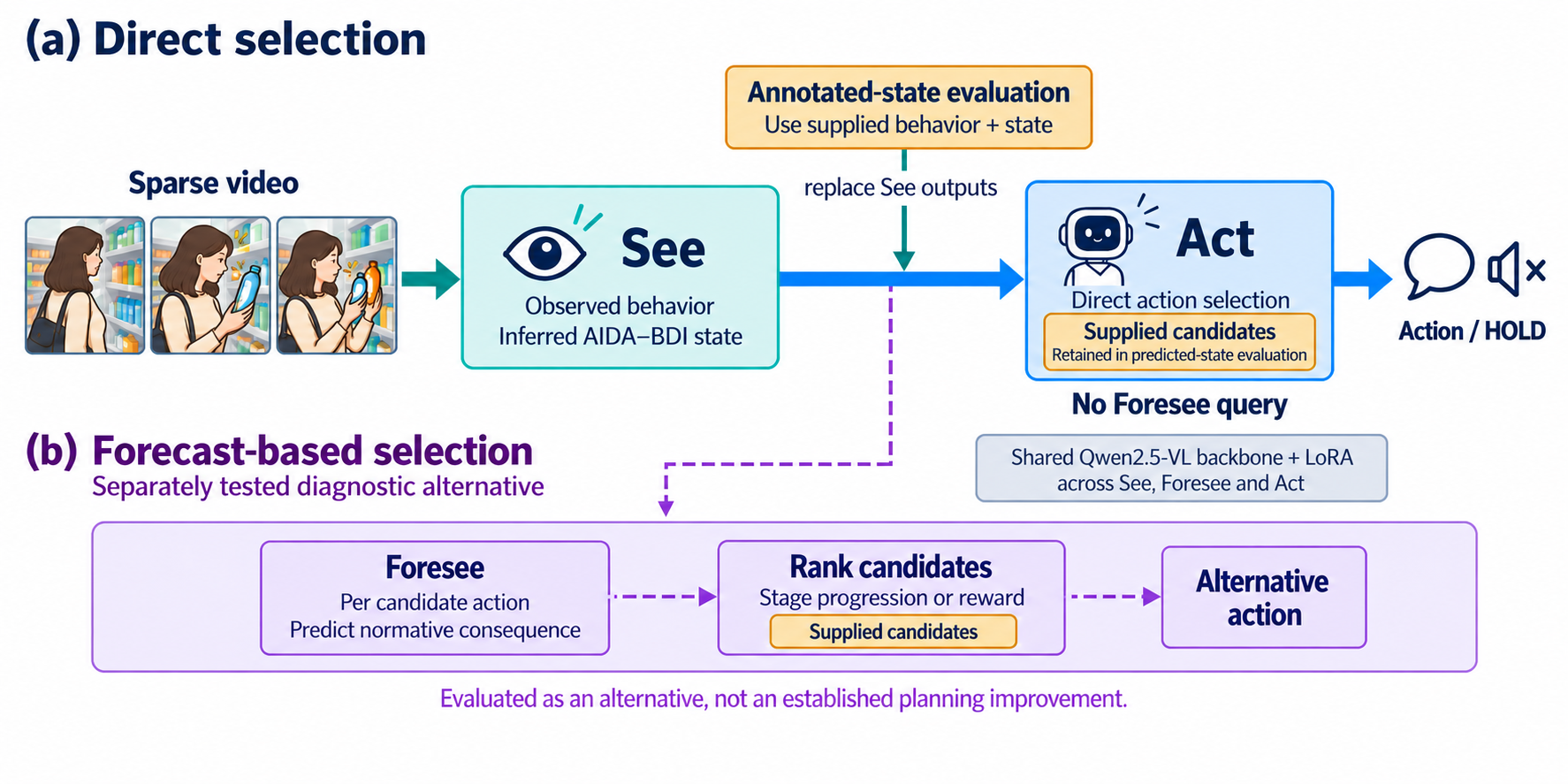}
\caption{See--Foresee--Act inference. See constructs a customer state from video; Foresee predicts one-step consequences for each candidate, including Hold; Act compares the candidate--consequence pairs to select an intervention. }\vspace{-15pt}
\label{fig:pipeline}
\end{figure}

\subsection{See: role-oriented evidence and customer
state}\label{see-role-oriented-evidence-and-customer-state}

See represents the current observation through explicit evidence and
inferred customer state. Four evidence fields,
\texttt{visible\_evidence}, \texttt{engagement\_pattern},
\texttt{gaze\_and\_attention}, and \texttt{body\_and\_hands}, describe
visible scene and behavior, engagement and interaction, attention, and
posture and hand movements. The retained state fields are
\texttt{stage}, \texttt{intent\_label}, \texttt{belief},
\texttt{desire}, and \texttt{intention}. AIDA describes task progress,
while BDI organizes beliefs, desires, and intentions. The inferred
fields can also encode information derived from visual observations.

Consider a customer repeatedly comparing two products and inspecting
their labels. A generic description records the movements; role-oriented
reading organizes the comparison targets and information-seeking cues.
The state representation expresses a possible need to understand product
differences and provides a common starting point for evaluating
interventions. This running example illustrates the interfaces rather
than reporting an experimental prediction.

\subsection{Foresee: one-step consequences of candidate
actions}\label{foresee-one-step-consequences-of-candidate-actions}

Foresee receives the current evidence, customer state, and one candidate
action. It predicts the next AIDA--BDI state, anticipated benefits and
risks, and a relative reward. Each candidate is evaluated from the same
current state. Hold is evaluated alongside active interventions,
allowing the model to compare the consequences of acting and remaining
silent.

In the running example, Elicit may clarify the comparison criteria,
Inform may address a specific information gap, and an early Recommend
may precede a clear understanding of preferences. Hold represents
continued independent comparison. Foresee expresses these alternatives
as candidate-specific next-state and benefit--risk descriptions for Act
to compare. Field schemas and inference interfaces appear in Appendices~\ref{app:interfaces} and~\ref{app:implementation}.

\subsection{Act: explicit consequence comparison and
non-intervention}\label{act-explicit-consequence-comparison-and-non-intervention}

Act receives current evidence and state together with the
candidate--consequence pairs produced by Foresee. It compares how well
each action addresses the inferred need, how it may advance the
interaction, and what interruption risks it introduces, then generates a
candidate label and a service realization. In the example, the choice
between Elicit and Inform depends on whether the comparison need has
become sufficiently specific; Hold remains available when independent
exploration is appropriate.

Non-intervention is an explicit decision alternative whose predicted
consequence participates in the same comparison as service actions. This
makes the decision concern both how to help and whether help is
currently warranted. The implementation uses a single decision point and
one-step candidate predictions.

\subsection{Training and inference}\label{training-and-inference}

PIWM parameterizes the three prediction tasks with a
Qwen2.5-VL-7B-Instruct backbone and LoRA adaptation.
Rather than training a monolithic video-to-action predictor, PIWM
decomposes supervision according to the model-based decision process:
current-state estimation, action-conditioned transition prediction, and
comparative action selection.

The central world-modeling objective is carried by \emph{Foresee}.
Given the current structured representation and a candidate action,
Foresee is supervised to predict the corresponding one-step customer
state together with its anticipated benefit and risk. Training on these
state--action--outcome tuples therefore teaches PIWM an
action-conditioned transition model over the structured customer state,
rather than only a direct policy from observations to actions.
\emph{See} provides the grounded current-state estimate on which these
transitions are conditioned, while \emph{Act} learns to select among
candidate actions using their predicted outcomes.

At inference time, the learned components are composed without access
to annotated states or consequences. See first constructs the current
representation from the sparse video. Foresee then evaluates each
candidate action from this shared current state, including \textsc{Hold},
and generates its predicted one-step outcome. Act finally compares the
resulting candidate--outcome pairs and selects the intervention.
This composition turns the learned transition model into an explicit
model-based decision procedure: actions are selected according to their
predicted consequences rather than from the current state alone.

\section{GuidanceSalesBench}\label{guidancesalesbench}

\subsection{Construction and
supervision}\label{construction-and-supervision}

Construction proceeds through four linked stages. \textbf{Knowledge
distillation} extracts 27 compact principles and 78 source-linked
conditional rules from retail and consumer-behavior material; these
counts describe an internal rule ledger, not independently established
laws. \textbf{Scenario specification} instantiates product context,
viewpoint, visible cues, and AIDA--BDI fields in a scene manifest.
\textbf{Candidate annotation} pairs a parent state with alternative
service realizations and LLM-assisted, rule-guided consequence targets,
followed by the expert review described in the construction records.
\textbf{Task export} turns the same parent material into video-to-state
See records, state--action-to-consequence Foresee records, best-action
Act records, and predicted-state evaluation inputs.

\begin{table}[tbp]
\centering\small
\setlength{\tabcolsep}{4pt}
\renewcommand{\arraystretch}{1.15}
\begin{tabularx}{\linewidth}{@{}lrX@{}}
\toprule
Data resource & Records & Purpose \\
\midrule
See & 1,043 & Multimodal observation representation \\
Foresee & 2,001 & Future-state prediction \\
Act & 680 & Decisions from textual states \\
Full & 3,054 & Multimodal-input-to-decision training data \\
Real & 418 & Sales-expert-annotated real-store scenes; evaluation only \\
\bottomrule
\end{tabularx}
\caption{Task-specific data resources. Full is split into 3,054 training, 509 validation, and 509 test examples. Real is excluded from training.}
\label{tab:results-2}
\end{table}

\subsection{Dataset composition and splits}
\label{dataset-composition}

\textsc{GuidanceSalesBench} contains task-specific resources for
state estimation, future-state prediction, and action selection,
together with an end-to-end multimodal collection.
Table~\ref{tab:results-2} summarizes the corresponding data
resources.

The component resources use different supervision units.
See contains 1,043 multimodal observation records, Foresee contains
2,001 state--action records, and Act contains 680 textual-state
decision records. Because these resources are task-specific exports,
their record counts should not be interpreted as counts of independent
scenes or summed with the Full collection.

A separate Real set contains 418 sales-expert-annotated real-store
records. These examples are excluded from training and reserved for
evaluation.

\subsection{A complete annotation
example}\label{a-complete-annotation-example}

\begin{figure*}[t]
\centering
\includegraphics[width=\textwidth]{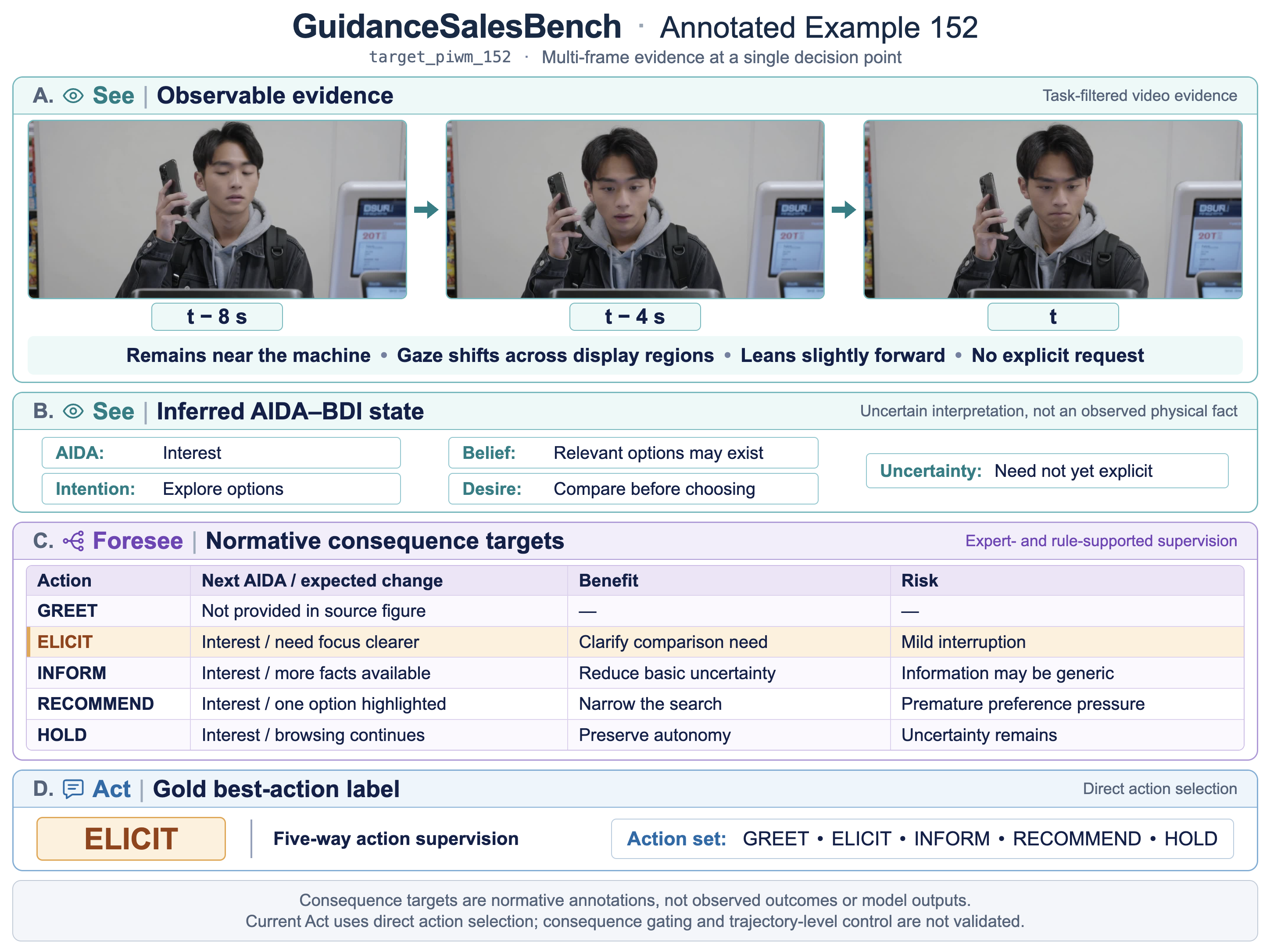}
\caption{A complete \textsc{GuidanceSalesBench} sample. Observed video evidence supports \emph{See}; the AIDA--BDI state is an inferred interpretation; each candidate action is paired with an expert- and rule-supported normative consequence target for \emph{Foresee}; and the gold best-action label supplies \emph{Act} supervision.}
\label{fig:guidancesalesbench-sample}
\end{figure*}

Figure~\ref{fig:guidancesalesbench-sample} illustrates a complete GuidanceSalesBench example across the four supervision layers. The sparse video frames show a customer remaining near the terminal, shifting gaze across display regions, and leaning slightly forward, without making an explicit request.

See records these observable cues and infers an AIDA \textit{Interest} state, together with the belief that relevant options may exist, the desire to compare before choosing, and the intention to explore available options. These inferred fields are task interpretations rather than directly observed facts.

Given this state, Foresee associates each candidate action with a normative consequence target. Elicit is annotated as clarifying the customer's comparison need with mild interruption risk; Inform may reduce basic uncertainty but can remain generic; Recommend may narrow the search but risks premature preference pressure; and Hold preserves independent exploration while leaving uncertainty unresolved. The preferred action is therefore Elicit.

The example illustrates how a single parent observation is transformed into observable evidence, an inferred customer state, candidate-specific consequence targets, and a gold best-action label. The consequence targets are normative annotations rather than outcomes observed after real interventions.

\section{Experiments}\label{experiments}

\subsection{Experimental setup}\label{experimental-setup}

We evaluate PIWM at both the component and end-to-end levels.
The component evaluations isolate the three structured prediction tasks
defined in Section~\ref{method}. \emph{See} is evaluated on 209
held-out examples and is counted as correct only when both the predicted
AIDA stage and intent label match the reference annotation.
\emph{Foresee} evaluates action-conditioned next-state prediction on
309 state--action records, using next-AIDA-stage accuracy over four
classes. \emph{Act} evaluates intervention selection on 245 examples
where the reference customer state is provided.

The \emph{Full} evaluation measures the complete video-to-action
pipeline on the 509-example held-out test split described in
Section~\ref{dataset-composition}. Unlike the component evaluations,
Full begins from sparse visual observations: See first predicts the
current customer representation, Foresee generates candidate-specific
outcomes when enabled, and Act produces the final intervention.
Consequently, Full captures the effect of composing the learned
components rather than evaluating any one prediction task in isolation.
Invalid or missing predictions are counted as errors.

For all tasks, we report accuracy:
\[
\mathrm{Accuracy}
=
\frac{1}{N}
\sum_{i=1}^{N}
\mathbf{1}[\hat y_i=y_i^\star].
\]
The theoretical random baselines follow the corresponding label spaces:
3.13\% for the joint four-stage/eight-intent See target, 25.00\% for
the four-class Foresee target, and 20.00\% for the five-action Act and
Full decisions.

\subsection{Overall performance}\label{overall-performance}

\begin{table}[tbp]
\centering\small
\setlength{\tabcolsep}{4pt}
\renewcommand{\arraystretch}{1.15}
\begin{tabularx}{\linewidth}{@{}Xrrrr@{}}
\toprule
Model & See (209) & Foresee (309) & Act (245) & Full (509) \\
\midrule
Random Baseline & 3.13 & 25.00 & 20.00 & 20.00 \\
Qwen2.5-VL-7B & 43.54 & 9.39 & 13.47 & 24.75 \\
Gemini 2.5 Flash & --- & --- & --- & \textbf{69.74} \\
GPT-4o & --- & --- & --- & 68.96 \\
PIWM (direct decision) &
\textbf{62.20} & \textbf{74.43} & \textbf{89.80} & 45.38 \\
PIWM (counterfactual) &
\textbf{62.20} & \textbf{74.43} & \textbf{89.80} & 63.65 \\
\bottomrule
\end{tabularx}
\caption{Component and full-pipeline accuracy (\%). The two PIWM
settings share the same See, Foresee, and Act evaluations and differ
only in Full inference: direct decision selects from the predicted
current representation without using Foresee outputs, whereas the
counterfactual setting supplies predicted candidate consequences to
Act. Bold indicates the highest reported score in each column.}
\label{tab:results-4}
\end{table}

PIWM improves substantially over the untuned Qwen2.5-VL-7B backbone on
all three component tasks. See increases from 43.54\% (91/209) to
62.20\% (130/209), indicating that the adapted model more reliably
recovers the joint AIDA-stage and intent representation from sparse
visual observations. Foresee increases from 9.39\% (29/309) to
74.43\% (230/309). This task directly evaluates the
state--action-conditioned next-state prediction used by the
world-modeling component of PIWM. Act reaches 89.80\% (220/245),
compared with 13.47\% (33/245) for the base model, when the structured
current state is provided.

The end-to-end Full evaluation tests whether these learned components
remain useful when composed from video input. Qwen2.5-VL-7B obtains
24.75\% accuracy (126/509). PIWM reaches 45.38\% (231/509) when Act
selects directly from the predicted current representation without
receiving explicit candidate forecasts. When Foresee predictions for
the candidate actions are supplied to Act, accuracy rises to 63.65\%
(324/509), corresponding to 93 additional correct decisions and an
improvement of 18.27 percentage points. GPT-4o and Gemini 2.5 Flash achieve 68.96\% (351/509) and 69.74\%
(355/509), respectively, on Full.

\subsection{Ablation analysis}\label{ablation-analysis}

We further examine three design choices in the composed pipeline:
explicit comparison of candidate consequences, retention of observable
evidence alongside the inferred customer state, and explicit prediction
of the consequence of non-intervention. All ablations use the same
509-example Full test set and preserve the five-action decision space.
Table~\ref{tab:results-5} summarizes the results.

\begin{table}[tbp]
\centering
\small
\setlength{\tabcolsep}{4pt}
\renewcommand{\arraystretch}{1.15}
\begin{tabular}{
@{}
>{\raggedright\arraybackslash}p{(\linewidth - 6\tabcolsep) * \real{0.46}}
>{\raggedright\arraybackslash}p{(\linewidth - 6\tabcolsep) * \real{0.17}}
>{\raggedright\arraybackslash}p{(\linewidth - 6\tabcolsep) * \real{0.17}}
>{\raggedright\arraybackslash}p{(\linewidth - 6\tabcolsep) * \real{0.2}}
@{}}
\toprule\noalign{}
Inference setting &
Correct / total &
Decision accuracy &
Difference from full pipeline \\
\midrule\noalign{}
Predicted state + Foresee + Act &
324 / 509 & 63.65\% & --- \\
Predicted state $\rightarrow$ Act directly &
231 / 509 & 45.38\% & $-18.27$ pp \\
Without explicit evidence fields &
141 / 509 & 27.70\% & $-35.95$ pp \\
Hold forecast unavailable &
244 / 509 & 47.94\% & $-15.72$ pp \\
\bottomrule
\end{tabular}
\caption{Ablations on the 509-example Full test set. Differences are
measured against the complete predicted-state + Foresee + Act pipeline.}
\label{tab:results-5}
\vspace{-15pt}
\end{table}

\paragraph{Consequence comparison.}
The complete pipeline predicts the current customer representation,
generates one-step consequences for the five candidate actions, and
passes the resulting candidate--consequence pairs to Act. The
direct-decision condition removes these forecasts and instead asks Act
to select an intervention directly from the predicted current state.
Accuracy decreases from 63.65\% to 45.38\%, a drop of 18.27 percentage
points and 93 correct decisions. Because the underlying PIWM components
and final action space are unchanged, this comparison isolates the
decision-time contribution of explicitly using action-conditioned
forecasts. The result supports the central premise of See--Foresee--Act:
predicting future states is useful when those predictions participate
in the final action comparison.

\paragraph{Explicit evidence.}
The evidence ablation removes
\texttt{visible\_evidence},
\texttt{engagement\_pattern},
\texttt{gaze\_and\_attention}, and
\texttt{body\_and\_hands} from the predicted current representation,
while retaining \texttt{stage}, \texttt{intent\_label},
\texttt{belief}, \texttt{desire}, and \texttt{intention}.
Foresee then regenerates all candidate consequences from this reduced
representation before Act performs the final comparison. Accuracy falls
to 27.70\% (141/509), 35.95 percentage points below the full pipeline
and 183 fewer correct decisions.

This ablation therefore measures the end-to-end contribution of
preserving explicit observable evidence rather than only whether Act
uses several additional text fields. The retained intent and BDI fields
may still contain information inferred from the visual input, so the
comparison does not remove all visually derived information. Instead,
it isolates the additional value of retaining explicit descriptions of
gaze, engagement, posture, and object interaction alongside the
structured customer state.

\paragraph{Non-intervention forecast.}
The Hold ablation retains all five candidate actions---Greet, Elicit,
Inform, Recommend, and Hold---together with the full predicted current
state and the explicit instruction that non-intervention is a valid
choice. The Hold forecast is withheld from Act, while forecasts for
the four active interventions remain available.

With the Hold forecast withheld, accuracy decreases to 47.94\%
(244/509), a drop of 15.72 percentage points and 80 correct decisions.
Because Hold remains selectable under both conditions, the comparison
distinguishes access to an anticipated outcome for waiting from merely
including Hold in the candidate set. Within the evaluated pipeline,
explicitly modeling the expected consequence of non-intervention
provides useful information for candidate comparison.

\section{Discussion: extending single-step decisions over
time}\label{discussion-extending-single-step-decisions-over-time}

PIWM connects task-oriented observation, one-step consequence
prediction, and explicit action comparison. Extending this formulation
to sustained interaction calls for tracking how new observations revise
customer state and how an earlier intervention changes later
opportunities. Multi-step trajectories and the value of waiting are
natural directions for extending the current single-decision framework.

 \vspace{-10pt}

\section{Limitations and Next Validation
Steps}\label{limitations-and-next-validation-steps}

The experiments study single-decision prediction and selection on a
509-example Full test set with synthetic, normative
supervision. The evidence ablation changes both Foresee and Act inputs
and regenerates forecasts, measuring their combined response to explicit
evidence removal. The Hold control measures the effect of a missing Hold
forecast while retaining the action and non-intervention instruction.
These results characterize decision accuracy within the evaluated
framework; longer interactions and customer outcomes require other
evaluation settings.

Future work can examine generalization across source scenes and
real-store interactions, and trace how state estimates and candidate
forecasts affect individual decisions. These extensions connect the
current single-step evaluation to sustained proactive assistance.

\section{Conclusion}\label{conclusion}

PIWM organizes proactive retail assistance through role-oriented state
construction, one-step consequence prediction, and explicit action
comparison. On the 509-example Full test set, consequence-based selection achieved 63.65\%
accuracy, compared with 45.38\% for direct decision output. Removing
explicit evidence fields reduced accuracy to 27.70\%; withholding only
the Hold forecast reduced it to 47.94\%. Together, these results support
preserving observable evidence and explicitly comparing the anticipated
consequences of intervention and non-intervention.

\section{AI Use Statement}\label{ai-use-statement}

Generative models were used in constructing synthetic data and normative
annotations. AI assistance was also used for literature discovery,
manuscript restructuring and drafting, interpretation of archived
experimental reports, and suggestions for additional experiments. 

\section{Ethics Statement}\label{ethics-statement}

Retail observations can expose personal behavior, and inferred intent
may be wrong. The real-store material in this study is staged; this work
does not evaluate covert deployment. Assistance should preserve the
option to decline and should not infer sensitive characteristics from
appearance. Data release requires verified consent and media rights.
Neither conformity to a sales-stage label nor an increase in inferred
purchase intention is a substitute for user welfare. The present
experiments do not establish customer benefit or deployment safety.

\section{Reproducibility Statement}\label{reproducibility-statement}

Appendix~\ref{app:data} describes the current resources and Full partition.
Appendices~\ref{app:interfaces} and~\ref{app:implementation} specify the
structured interfaces and inference controls.
Appendix~\ref{app:evaluation} gives task-specific scoring definitions and
correct counts. Appendix~\ref{app:reproduction} describes the Real
evaluation coverage and anonymous-material version.

\bibliography{references_iclr}

@inproceedings{horvitz1999principles,
  title={Principles of mixed-initiative user interfaces},
  author={Horvitz, Eric},
  booktitle={Proceedings of the SIGCHI conference on Human Factors in Computing Systems},
  pages={159--166},
  year={1999}
}

@inproceedings{rabinowitz2018machine,
  title={Machine theory of mind},
  author={Rabinowitz, Neil and Perbet, Frank and Song, Francis and Zhang, Chiyuan and Eslami, SM Ali and Botvinick, Matthew},
  booktitle={International conference on machine learning},
  pages={4218--4227},
  year={2018},
  organization={PMLR}
}

@inproceedings{rao1995bdi,
  title={{BDI} Agents: From Theory to Practice},
  author={Rao, Anand S and Georgeff, Michael P},
  booktitle={Icmas},
  volume={95},
  pages={312--319},
  year={1995}
}

@article{barry1987development,
  title={The development of the hierarchy of effects: An historical perspective},
  author={Barry, Thomas E},
  journal={Current issues and Research in Advertising},
  volume={10},
  number={1-2},
  pages={251--295},
  year={1987},
  publisher={Taylor \& Francis}
}

@article{ha2018world,
  title={{World Models}},
  author={Ha, David and Schmidhuber, J{\"u}rgen},
  journal={arXiv preprint arXiv:1803.10122},
  year={2018},
  url={https://arxiv.org/abs/1803.10122}
}

@inproceedings{hafner2020dreamer,
  title={Dream to Control: Learning Behaviors by Latent Imagination},
  author={Hafner, Danijar and Lillicrap, Timothy and Ba, Jimmy and Norouzi, Mohammad},
  booktitle={International Conference on Learning Representations},
  year={2020},
  url={https://openreview.net/forum?id=S1lOTC4tDS}
}

@misc{bai2025qwen25vltechnicalreport,
      title={{Qwen2.5-VL} Technical Report}, 
      author={Shuai Bai and Keqin Chen and Xuejing Liu and Jialin Wang and Wenbin Ge and Sibo Song and Kai Dang and Peng Wang and Shijie Wang and Jun Tang and Humen Zhong and Yuanzhi Zhu and Mingkun Yang and Zhaohai Li and Jianqiang Wan and Pengfei Wang and Wei Ding and Zheren Fu and Yiheng Xu and Jiabo Ye and Xi Zhang and Tianbao Xie and Zesen Cheng and Hang Zhang and Zhibo Yang and Haiyang Xu and Junyang Lin},
      year={2025},
      eprint={2502.13923},
      archivePrefix={arXiv},
      primaryClass={cs.CV},
      url={https://arxiv.org/abs/2502.13923}, 
}

@article{yang2026contextagent,
  title={{ContextAgent}: Context-Aware Proactive {LLM} Agents with Open-World Sensory Perceptions},
  author={Yang, Bufang and Xu, Lilin and Zeng, Liekang and Liu, Kaiwei and Jiang, Siyang and Lu, Wenrui and Chen, Hongkai and Jiang, Xiaofan and Xing, Guoliang and Yan, Zhenyu},
  journal={Advances in Neural Information Processing Systems},
  volume={38},
  pages={167509--167543},
  year={2025}
}

@article{fu2026prism,
  title={{PRISM}: Festina Lente Proactivity---Risk-Sensitive, Uncertainty-Aware Deliberation for Proactive Agents},
  author={Fu, Yuxuan and Tan, Xiaoyu and Hao, Teqi and Zhan, Chen and Qiu, Xihe},
  journal={arXiv preprint arXiv:2602.01532},
  year={2026},
  url={https://arxiv.org/abs/2602.01532}
}

@article{lu2024proactive,
 title={Proactive Agent: Shifting {LLM} Agents from Reactive Responses to Active Assistance},
 author={Lu, Yaxi and Yang, Shenzhi and Qian, Cheng and Chen, Guirong and Luo, Qinyu and Wu, Yesai and Wang, Huadong and Cong, Xin and Zhang, Zhong and Lin, Yankai and Liu, Weiwen and Wang, Yasheng and Liu, Zhiyuan and Liu, Fangming and Sun, Maosong},
 journal={arXiv preprint arXiv:2410.12361},year={2024},url={https://arxiv.org/abs/2410.12361}
}

@inproceedings{chen2025llamapie,
 title={{LlamaPIE}: Proactive In-Ear Conversation Assistants},
 author={Chen, Tuochao and Batchelder, Nicholas Scott and Liu, Alisa and Smith, Noah A. and Gollakota, Shyamnath},
 booktitle={Findings of the Association for Computational Linguistics: ACL 2025},
 year={2025},pages={13801--13824},doi={10.18653/v1/2025.findings-acl.710},url={https://aclanthology.org/2025.findings-acl.710/}
}

@inproceedings{zhang2025proassist,
 title={Proactive Assistant Dialogue Generation from Streaming Egocentric Videos},
 author={Zhang, Yichi and Dong, Xin Luna and Lin, Zhaojiang and Madotto, Andrea and Kumar, Anuj and Damavandi, Babak and Chai, Joyce and Moon, Seungwhan},
 booktitle={Proceedings of the 2025 Conference on Empirical Methods in Natural Language Processing},
 year={2025},pages={12044--12068},doi={10.18653/v1/2025.emnlp-main.605},url={https://aclanthology.org/2025.emnlp-main.605/}
}

@inproceedings{xiang2024language,
  title={Language Models Meet World Models: Embodied Experiences Enhance Language Models},
  author={Xiang, Jiannan and Tao, Tianhua and Gu, Yi and Shu, Tianmin and Wang, Zirui and Yang, Zichao and Hu, Zhiting},
  booktitle={Advances in Neural Information Processing Systems},
  volume={36},
  pages={75392--75412},
  publisher={Curran Associates, Inc.},
  doi={10.52202/075280-3295},
  url={https://proceedings.neurips.cc/paper_files/paper/2023/hash/ee6630dcbcff857026e474fc857aa9f0-Abstract-Conference.html},
  year={2023}
}

@inproceedings{shridhar2020alfred,
  title     = {{ALFWorld}: Aligning Text and Embodied Environments for Interactive Learning},
  author    = {Shridhar, Mohit and Yuan, Xingdi and C{\^o}t{\'e}, Marc-Alexandre and Bisk, Yonatan and Trischler, Adam and Hausknecht, Matthew},
  booktitle = {International Conference on Learning Representations},
  year      = {2021},
  url       = {https://openreview.net/forum?id=0IOX0YcCdTn}
}

@inproceedings{Wang2022ScienceWorld,
    title = "{S}cience{W}orld: Is your Agent Smarter than a 5th Grader?",
    author = "Wang, Ruoyao  and
      Jansen, Peter  and
      C{\^o}t{\'e}, Marc-Alexandre  and
      Ammanabrolu, Prithviraj",
    booktitle = "Proceedings of the 2022 Conference on Empirical Methods in Natural Language Processing",
    month = dec,
    year = "2022",
    address = "Abu Dhabi, United Arab Emirates",
    publisher = "Association for Computational Linguistics",
    pages = "11279--11298",
    doi = "10.18653/v1/2022.emnlp-main.775",
    url = "https://aclanthology.org/2022.emnlp-main.775/",
}

@inproceedings{yao2023react,
  title = {{ReAct}: Synergizing Reasoning and Acting in Language Models},
  author = {Yao, Shunyu and Zhao, Jeffrey and Yu, Dian and Du, Nan and Shafran, Izhak and Narasimhan, Karthik and Cao, Yuan},
  booktitle = {The Eleventh International Conference on Learning Representations},
  year = {2023},
  url = {https://openreview.net/forum?id=WE_vluYUL-X},
}

@inproceedings{wang-etal-2025-steca,
    title = "{ST}e{C}a: Step-level Trajectory Calibration for {LLM} Agent Learning",
    author = "Wang, Hanlin  and
      Wang, Jian  and
      Leong, Chak Tou  and
      Li, Wenjie",
    editor = "Che, Wanxiang  and
      Nabende, Joyce  and
      Shutova, Ekaterina  and
      Pilehvar, Mohammad Taher",
    booktitle = "Findings of the Association for Computational Linguistics: ACL 2025",
    month = jul,
    year = "2025",
    address = "Vienna, Austria",
    publisher = "Association for Computational Linguistics",
    url = "https://aclanthology.org/2025.findings-acl.604/",
    doi = "10.18653/v1/2025.findings-acl.604",
    pages = "11597--11614",
}

@inproceedings{song-etal-2024-trial,
    title = "Trial and Error: Exploration-Based Trajectory Optimization of {LLM} Agents",
    author = "Song, Yifan  and
      Yin, Da  and
      Yue, Xiang  and
      Huang, Jie  and
      Li, Sujian  and
      Lin, Bill Yuchen",
    editor = "Ku, Lun-Wei  and
      Martins, Andre  and
      Srikumar, Vivek",
    booktitle = "Proceedings of the 62nd Annual Meeting of the Association for Computational Linguistics (Volume 1: Long Papers)",
    month = aug,
    year = "2024",
    address = "Bangkok, Thailand",
    publisher = "Association for Computational Linguistics",
    url = "https://aclanthology.org/2024.acl-long.409/",
    doi = "10.18653/v1/2024.acl-long.409",
    pages = "7584--7600"
}

@inproceedings{xiong-etal-2024-watch,
    title = "Watch Every Step! {LLM} Agent Learning via Iterative Step-level Process Refinement",
    author = "Xiong, Weimin  and
      Song, Yifan  and
      Zhao, Xiutian  and
      Wu, Wenhao  and
      Wang, Xun  and
      Wang, Ke  and
      Li, Cheng  and
      Peng, Wei  and
      Li, Sujian",
    editor = "Al-Onaizan, Yaser  and
      Bansal, Mohit  and
      Chen, Yun-Nung",
    booktitle = "Proceedings of the 2024 Conference on Empirical Methods in Natural Language Processing",
    month = nov,
    year = "2024",
    address = "Miami, Florida, USA",
    publisher = "Association for Computational Linguistics",
    url = "https://aclanthology.org/2024.emnlp-main.93/",
    doi = "10.18653/v1/2024.emnlp-main.93",
    pages = "1556--1572"
}

@inproceedings{puig2018virtualhome,
  title={{VirtualHome}: Simulating Household Activities via Programs},
  author={Puig, Xavier and Ra, Kevin and Boben, Marko and Li, Jiaman and Wang, Tingwu and Fidler, Sanja and Torralba, Antonio},
  booktitle={Proceedings of the IEEE conference on computer vision and pattern recognition},
  pages={8494--8502},
  year={2018}
}

@inproceedings{shinn2023reflexion,
  title={Reflexion: Language agents with verbal reinforcement learning},
  author={Shinn, Noah and Cassano, Federico and Gopinath, Ashwin and Narasimhan, Karthik and Yao, Shunyu},
  booktitle={Advances in Neural Information Processing Systems},
  volume={36},
  pages={8634--8652},
  year={2023}
}

@article{hafner2023mastering,
  title={Mastering diverse domains through world models},
  author={Hafner, Danijar and Pasukonis, Jurgis and Ba, Jimmy and Lillicrap, Timothy},
  journal={arXiv preprint arXiv:2301.04104},
  year={2023}
}

@inproceedings{wang-etal-2025-world,
    title = "World Modeling Makes a Better Planner: Dual Preference Optimization for Embodied Task Planning",
    author = "Wang, Siyin  and
      Fei, Zhaoye  and
      Cheng, Qinyuan  and
      Zhang, Shiduo  and
      Cai, Panpan  and
      Fu, Jinlan  and
      Qiu, Xipeng",
    editor = "Che, Wanxiang  and
      Nabende, Joyce  and
      Shutova, Ekaterina  and
      Pilehvar, Mohammad Taher",
    booktitle = "Proceedings of the 63rd Annual Meeting of the Association for Computational Linguistics (Volume 1: Long Papers)",
    month = jul,
    year = "2025",
    address = "Vienna, Austria",
    publisher = "Association for Computational Linguistics",
    url = "https://aclanthology.org/2025.acl-long.1044/",
    doi = "10.18653/v1/2025.acl-long.1044",
    pages = "21518--21537",
}

@inproceedings{hao-etal-2023-reasoning,
    title = "Reasoning with Language Model is Planning with World Model",
    author = "Hao, Shibo  and
      Gu, Yi  and
      Ma, Haodi  and
      Hong, Joshua  and
      Wang, Zhen  and
      Wang, Daisy  and
      Hu, Zhiting",
    editor = "Bouamor, Houda  and
      Pino, Juan  and
      Bali, Kalika",
    booktitle = "Proceedings of the 2023 Conference on Empirical Methods in Natural Language Processing",
    month = dec,
    year = "2023",
    address = "Singapore",
    publisher = "Association for Computational Linguistics",
    url = "https://aclanthology.org/2023.emnlp-main.507/",
    doi = "10.18653/v1/2023.emnlp-main.507",
    pages = "8154--8173",
}

@inproceedings{fang-etal-2025-webevolver,
    title = "{W}eb{E}volver: Enhancing Web Agent Self-Improvement with Co-evolving World Model",
    author = "Fang, Tianqing  and
      Zhang, Hongming  and
      Zhang, Zhisong  and
      Ma, Kaixin  and
      Yu, Wenhao  and
      Mi, Haitao  and
      Yu, Dong",
    editor = "Christodoulopoulos, Christos  and
      Chakraborty, Tanmoy  and
      Rose, Carolyn  and
      Peng, Violet",
    booktitle = "Proceedings of the 2025 Conference on Empirical Methods in Natural Language Processing",
    month = nov,
    year = "2025",
    address = "Suzhou, China",
    publisher = "Association for Computational Linguistics",
    url = "https://aclanthology.org/2025.emnlp-main.454/",
    doi = "10.18653/v1/2025.emnlp-main.454",
    pages = "8959--8975",
}

@inproceedings{qiao2024agent,
  title={Agent planning with world knowledge model},
  author={Qiao, Shuofei and Fang, Runnan and Zhang, Ningyu and Zhu, Yuqi and Chen, Xiang and Deng, Shumin and Jiang, Yong and Xie, Pengjun and Huang, Fei and Chen, Huajun},
  booktitle={Advances in Neural Information Processing Systems},
  volume={37},
  pages={114843--114871},
  year={2024}
}

@inproceedings{guan2023leveraging,
  title={Leveraging pre-trained large language models to construct and utilize world models for model-based task planning},
  author={Guan, Lin and Valmeekam, Karthik and Sreedharan, Sarath and Kambhampati, Subbarao},
  booktitle={Advances in Neural Information Processing Systems},
  volume={36},
  pages={79081--79094},
  year={2023}
}

@inproceedings{wang2025vagen,
  title={{VAGEN}: Reinforcing World Model Reasoning for Multi-Turn {VLM} Agents},
  author={Wang, Kangrui and Zhang, Pingyue and Wang, Zihan and Gao, Yaning and Li, Linjie and Wang, Qineng and Chen, Hanyang and Lu, Yiping and Yang, Zhengyuan and Wang, Lijuan and Krishna, Ranjay and Wu, Jiajun and Li, Fei-Fei and Choi, Yejin and Li, Manling},
  booktitle={Advances in Neural Information Processing Systems},
  volume={38},
  pages={172871--172933},
  publisher={Curran Associates, Inc.},
  doi={10.52202/085713-5754},
  url={https://proceedings.neurips.cc/paper_files/paper/2025/hash/fc6688d75adde86b9df910769c1d02e3-Abstract-Conference.html},
  year={2025}
}

@article{zhang2025world,
  title={{World-in-World}: World Models in a Closed-Loop World},
  author={Zhang, Jiahan and Jiang, Muqing and Dai, Nanru and Lu, Taiming and Uzunoglu, Arda and Zhang, Shunchi and Wei, Yana and Wang, Jiahao and Patel, Vishal M and Liang, Paul Pu and others},
  journal={arXiv preprint arXiv:2510.18135},
  year={2025}
}

@article{aastrom1965optimal,
  title={Optimal Control of Markov Processes with Incomplete State Information {I}},
  author={{\AA}str{\"o}m, Karl Johan},
  journal={Journal of Mathematical Analysis and Applications},
  volume={10},
  pages={174--205},
  year={1965},
  publisher={Elsevier},
  doi={10.1016/0022-247X(65)90154-X},
  url={https://doi.org/10.1016/0022-247X(65)90154-X}
}

@inproceedings{yao2022webshop,
  author    = {Shunyu Yao and Howard Chen and John Yang and Karthik Narasimhan},
  title     = {{WebShop}: Towards Scalable Real-World Web Interaction with Grounded Language Agents},
  booktitle = {Advances in Neural Information Processing Systems},
  volume    = {35},
  pages     = {20744--20757},
  year      = {2022}
}
\bibliographystyle{iclr2027_conference}
\clearpage
\appendix
\section{Data resources, review, and partitioning}\label{app:data}

\subsection{Resource units and evaluation
sets}\label{resource-units-and-evaluation-sets}

GuidanceSalesBench organizes supervision around observations, customer
states, candidate consequences, and service decisions. See evaluates
multimodal observation representation; Foresee evaluates next-state
prediction; Act evaluates decisions conditioned on textual reference
states; Full evaluates the complete multimodal-input-to-action pipeline.
Real supplies sales-expert-annotated records from staged real-store
scenes.

\begin{table}[!htbp]
\centering\small
\setlength{\tabcolsep}{4pt}
\renewcommand{\arraystretch}{1.15}
\begin{tabularx}{\linewidth}{@{}lrrX@{}}
\toprule
Resource & Resource records & Evaluation records & Evaluation target \\

\midrule

See & 1,043 & 209 & Jointly correct AIDA stage and intent \\
Foresee & 2,001 & 309 & Next AIDA stage \\
Act & 680 & 245 & Selected action category \\
Full & 3,054 training & 509 test & Final video-to-action decision \\
Real & 418 & 418 & Real-store action choice \\

\bottomrule
\end{tabularx}
\caption{Current resource and evaluation counts. Full additionally has
509 validation examples. Real is reserved exclusively for
evaluation.}
\end{table}

The reviewed Full pool contains 4,072 examples retained from an initial
pool of 6,000 synthetic examples after review by large language models
and sales experts. Its partition is 3,054 training, 509 validation, and
509 test examples, corresponding to a 6:1:1 ratio. The Full results and
inference-time ablations in this paper use the 509-example test set. The
validation split is distinct from the reported test split. Real records
do not participate in training.

The component counts refer to task-specific resources and their
corresponding evaluation sets. A Foresee record is a state--action pair,
whereas a Full record covers a complete observation-to-decision example.
The Full split ratio does not define the component splits.

\subsection{Annotation structure}\label{annotation-structure}

Each example connects four levels: a pre-interaction observation, a
structured current state, candidate-conditioned consequences, and a
preferred service action. The current state contains explicit observable
evidence together with an inferred AIDA stage, intent label, and BDI
description. Consequence annotations specify the next state and the
anticipated benefit, risk, and reward associated with a candidate
action.

A consequence target describes the expected one-step outcome under the
benchmark's retail-service interpretation. The preferred action is
selected within that same interpretation. This connects state estimation
and action-conditioned prediction to the decision label used for
evaluation.

\subsection{Real-store evaluation}\label{real-store-evaluation}


Real contains 418 sales-expert-annotated records from staged real-store
scenes. It is used to examine service decisions under real-store visual
observations. The evaluation target is the selected service action. Its
records are kept separate from synthetic training data and from the Full
test set.

\section{Structured interfaces and a running
example}\label{app:interfaces}

\subsection{Current-state
representation}\label{current-state-representation}

The state interface separates observable evidence from inferred task
state.

\begin{table}[!htbp]
\centering\small
\setlength{\tabcolsep}{4pt}
\renewcommand{\arraystretch}{1.15}
\begin{tabularx}{\linewidth}{@{}p{0.17\linewidth}p{0.43\linewidth}X@{}}
\toprule
Field group & Fields & Function \\

\midrule

Observable evidence & visible\_evidence, engagement\_pattern,
gaze\_and\_attention, body\_and\_hands & Describe the scene, engagement,
attention, posture, and hand movements \\
Task labels & stage, intent\_label & Represent purchasing stage and
intent category \\
BDI interpretation & belief, desire, intention & Express the inferred
customer state \\

\bottomrule
\end{tabularx}
\caption{Current-state fields supplied to downstream prediction and
decision modules.}
\end{table}

See predicts four AIDA stages and eight intent categories. Its
evaluation uses joint correctness of the two categorical fields. The
evidence and BDI fields remain part of the downstream representation
even though the See score is defined over stage and intent.

The interface can be serialized as follows:

{\small
\begin{verbatim}
<stage>...</stage>
<intent_label>...</intent_label>
<visible_evidence>...</visible_evidence>
<engagement_pattern>...</engagement_pattern>
<gaze_and_attention>...</gaze_and_attention>
<body_and_hands>...</body_and_hands>
<belief>...</belief>
<desire>...</desire>
<intention>...</intention>
\end{verbatim}
}

\subsection{Candidate consequences and action
output}\label{candidate-consequences-and-action-output}

Foresee takes the current state and one candidate action. Its output
specifies the next AIDA stage, next BDI state, risk, benefit, and
relative reward. Risk and benefit use low, medium, and high categories;
reward is represented on the interval {[}-1, 1{]}. Each of the five
service actions, including Hold, has its own consequence record.

{\small
\begin{verbatim}
<next_stage>...</next_stage>
<next_belief>...</next_belief>
<next_desire>...</next_desire>
<next_intention>...</next_intention>
<risk>low|medium|high</risk>
<benefit>low|medium|high</benefit>
<reward>...</reward>
\end{verbatim}
}

Act receives the current state and candidate--consequence pairs. Its
output includes a rationale, selected candidate, action realization, and
intervention utterance.

{\small
\begin{verbatim}
<rationale>...</rationale>
<chosen>...</chosen>
<intervention_action>...</intervention_action>
<intervention_utterance>...</intervention_utterance>
\end{verbatim}
}

These blocks specify output interfaces. See, Foresee, and Act have
different input requirements: See receives multimodal observations,
Foresee receives a state--action pair, and comparative Act receives the
current state and candidate forecasts.

\subsection{A running example}\label{a-running-example}

Consider a customer who pauses at a terminal, looks across the display,
and compares available options. See records these visible cues and
infers an Interest stage with an exploratory intent. A possible BDI
interpretation is that relevant products may be available, the customer
wants to compare their suitability, and the immediate intention is to
continue exploring.

Foresee evaluates candidate interventions from this same current state.
Elicit may clarify comparison criteria; Inform may address a specific
information gap; Recommend may be premature before preferences become
clear; Hold allows independent exploration to continue. Act compares
these alternatives using the current evidence, inferred need, and
anticipated consequences. This example illustrates the interfaces and
does not assign a model success score.

A useful consequence need not change the AIDA label. An intervention can
clarify preferences while leaving the customer in the Interest stage.
Accordingly, Foresee's next-stage score and Full decision accuracy
measure different parts of the pipeline.

\section{Model configuration and single-step
inference}\label{app:implementation}

\subsection{Model and task interfaces}\label{model-and-task-interfaces}

PIWM uses a Qwen2.5-VL-7B-Instruct backbone with LoRA adaptation. The three tasks share the structured state and
action interfaces. See estimates the current state from multimodal
observations. Foresee learns action-conditioned one-step consequences.
Act performs action selection from the information provided by its
decision configuration.

The two PIWM rows in the main table share the See, Foresee, and Act
component evaluations. Their Full configurations differ in whether
predicted consequences are supplied to the decision step. The
direct-decision configuration does not invoke Foresee during Full
inference.

\subsection{Candidate comparison}\label{candidate-comparison}

The candidate action categories are Greet, Elicit, Inform, Recommend,
and Hold. The full pipeline first computes the current state, then
predicts the consequence of each candidate, and finally supplies the
state and candidate--consequence pairs to Act. Each forecast starts from
the same current state. Predictions are not recursively rolled forward
into a multi-step imagined trajectory.

The following pseudocode specifies the input transformations used by the
four configurations:

{\small
\begin{verbatim}
state = See(observations)
if condition == "direct":
    decision = Act(state, candidates, consequences=None)
else:
    if condition == "without_explicit_evidence":
        state = remove_four_evidence_fields(state)
    forecasts = {a: Foresee(state, a) for a in candidates}
    if condition == "hold_forecast_unavailable":
        forecasts["Hold"] = {"prediction_unavailable": True}
    decision = Act(state, candidates, forecasts)
\end{verbatim}
}

\subsection{Controlled input changes}\label{controlled-input-changes}

\begin{table}[!htbp]
\centering\small
\setlength{\tabcolsep}{4pt}
\renewcommand{\arraystretch}{1.15}
\begin{tabularx}{\linewidth}{@{}p{0.22\linewidth}XXX@{}}
\toprule
Configuration & State supplied downstream & Consequences supplied to Act & Hold selectable \\

\midrule

Full pipeline & Complete predicted state & Five predicted outcomes &
Yes \\
Direct decision & Complete predicted state & None & Yes \\
Without explicit evidence & Four evidence fields removed & Regenerated
from the reduced state & Yes \\
Hold forecast unavailable & Complete predicted state & Four outcomes and
a missing marker for Hold & Yes \\

\bottomrule
\end{tabularx}
\caption{Inputs of the four inference configurations.}
\end{table}

\begin{figure*}[t]
    \centering
    \includegraphics[width=\textwidth]{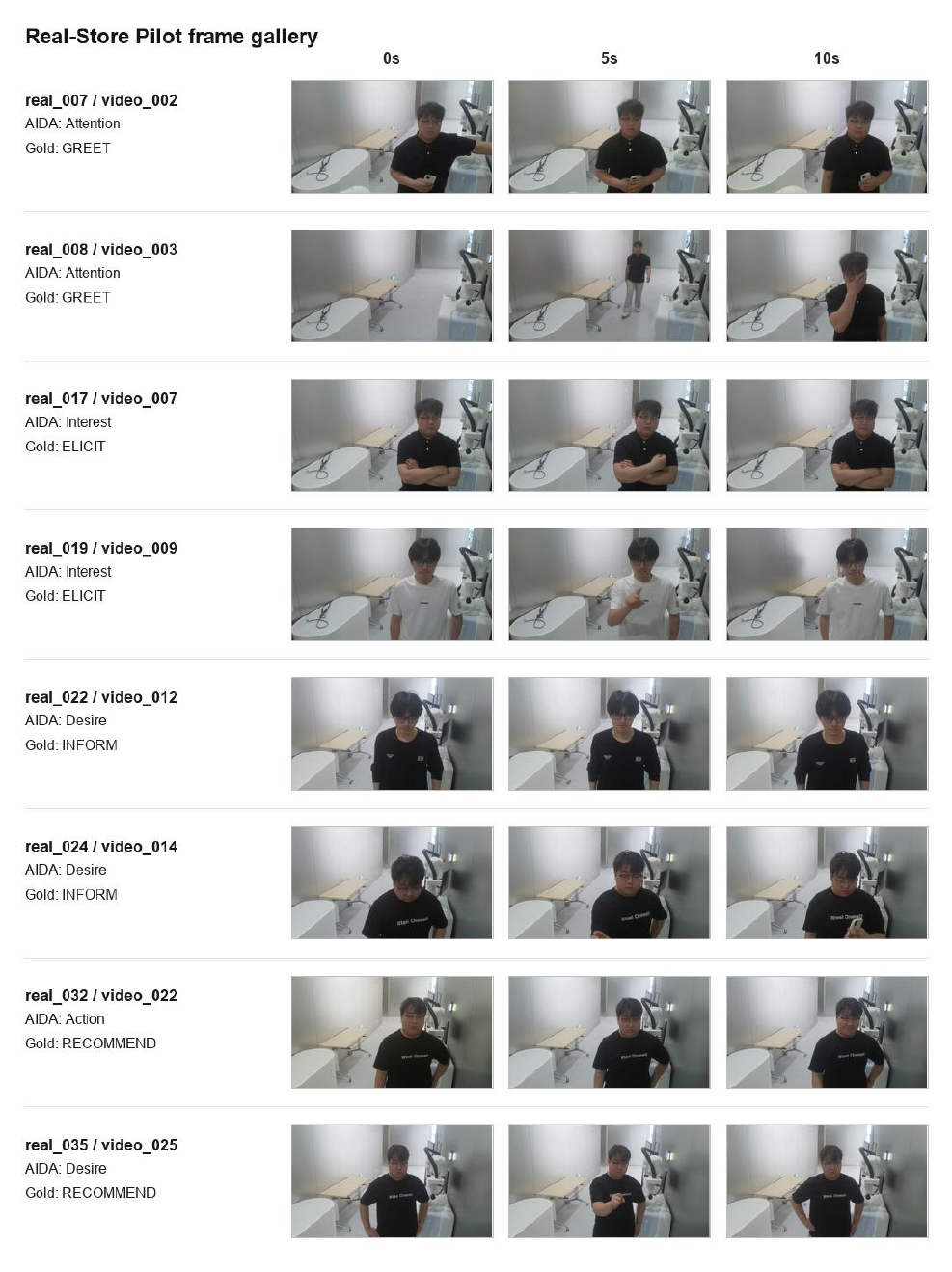}
    \caption{Additional frame examples from the real-store pilot.
    Each row shows a distinct paid-participant session, with rows
    spanning different gold-standard AIDA stages and best-action labels.
    All participants signed release forms authorizing public release
    of the recorded materials (see Ethics Statement).}
    \label{fig:real-store-gallery}
\end{figure*}

The evidence ablation removes visible\_evidence, engagement\_pattern,
gaze\_and\_attention, and body\_and\_hands. It retains stage,
intent\_label, belief, desire, and intention. Both Foresee and Act
receive the reduced state, and consequences are generated again from
that state. It therefore measures the combined downstream effect of
retaining explicit evidence.

The Hold ablation replaces only the Hold forecast with
\{``prediction\_unavailable'': true\}. The complete current state, other
four forecasts, five-action candidate set, and explicit non-intervention
instruction are retained. Hold remains a legal decision.

\section{Evaluation definitions and detailed
results}\label{app:evaluation}

\subsection{Task-specific accuracy}\label{task-specific-accuracy}

Let \(c_i\) denote whether example \(i\) is correctly predicted under
its task-specific target. Accuracy is

\[
\operatorname{Accuracy}=\frac{1}{N}\sum_{i=1}^{N}c_i.
\]

For See, stage and intent must both be correct:

\[
c_i^{\mathrm{See}}=
\mathbf{1}[\hat{s}_i=s_i]\,
\mathbf{1}[\hat{\ell}_i=\ell_i],
\qquad N_{\mathrm{See}}=209.
\]

A correct stage with an incorrect intent, or the reverse, receives zero
credit. Missing or invalid target fields also receive zero credit.

Foresee compares the predicted next AIDA stage with its reference over
309 state--action records. Act compares the chosen action category with
the reference over 245 textual reference-state examples. Full compares
the final action produced from multimodal observations with the
reference over 509 test examples. Real uses expert action labels over
its 418 evaluation records.

All records remain in the task denominator, including invalid or missing
predictions. Risk, benefit, reward, and generated BDI text are part of
the consequence representation; they are not additional terms in the
reported Foresee accuracy.

\subsection{Component results}\label{component-results}

\begin{table}[!htbp]
\centering\small
\setlength{\tabcolsep}{4pt}
\renewcommand{\arraystretch}{1.15}
\begin{tabularx}{\linewidth}{@{}Xrrr@{}}
\toprule
Model & See & Foresee & Act \\

\midrule

Qwen2.5-VL-7B & 91/209 (43.54\%) & 29/309 (9.39\%) & 33/245 (13.47\%) \\
PIWM & 130/209 (62.20\%) & 230/309 (74.43\%) & 220/245 (89.80\%) \\

\bottomrule
\end{tabularx}
\caption{Correct counts and accuracies for the current component
evaluations.}
\end{table}

The random baseline assumes uniform sampling of the corresponding
categorical targets. Its theoretical accuracies are 3.13\% for joint See
prediction, 25.00\% for Foresee, and 20.00\% for five-action decisions.

\subsection{Full results}\label{full-results}

\begin{table}[!htbp]
\centering\small
\setlength{\tabcolsep}{4pt}
\renewcommand{\arraystretch}{1.15}
\begin{tabularx}{\linewidth}{@{}Xrr@{}}
\toprule
Model or decision configuration & Correct / total & Accuracy \\

\midrule

Qwen2.5-VL-7B & 126 / 509 & 24.75\% \\
Gemini 2.5 Flash & 355 / 509 & 69.74\% \\
GPT-4o & 351 / 509 & 68.96\% \\
PIWM, direct decision & 231 / 509 & 45.38\% \\
PIWM, counterfactual reasoning & 324 / 509 & 63.65\% \\

\bottomrule
\end{tabularx}
\caption{Correct counts for the current Full test set.}
\end{table}

\subsection{Inference ablations}\label{inference-ablations}

\begin{table}[!htbp]
\centering\small
\setlength{\tabcolsep}{4pt}
\renewcommand{\arraystretch}{1.15}
\begin{tabularx}{\linewidth}{@{}Xrrr@{}}
\toprule
Configuration & Correct / total & Accuracy & Difference from full pipeline \\

\midrule

Direct decision & 231 / 509 & 45.38\% & \ensuremath{-}18.27 pp \\
Without explicit evidence & 141 / 509 & 27.70\% & \ensuremath{-}35.95 pp \\
Hold forecast unavailable & 244 / 509 & 47.94\% & \ensuremath{-}15.72 pp \\
Full pipeline & 324 / 509 & 63.65\% & --- \\

\bottomrule
\end{tabularx}
\caption{Inference ablations on the 509-example Full test
set.}
\end{table}

Consequence comparison adds 93 correct decisions relative to direct
output. Removing explicit evidence reduces the total by 183 correct
decisions; withholding the Hold forecast reduces it by 80. These are
aggregate differences over the same test-set size.

For paired correctness indicators, let \(b\) count examples correct only
under condition A and \(c\) examples correct only under condition B. The
accuracy difference is \((b-c)/N\). This distinction separates the net
difference in correct counts from the number of predictions that change
in each direction.

\section{Real-store material and
reproducibility}\label{app:reproduction}

\subsection{Real-store observations}\label{real-store-observations}

The Real collection uses staged retail interactions and sales-expert
action annotations. Its 418 records are reserved for evaluation. The
task remains a single-decision service choice from the observed customer
context, using the same five action categories as the synthetic setting.

Real-store material can contain identifiable people and background
information. The scope of research access, paper illustration, and
public media redistribution is handled separately. The main Ethics
Statement describes the conditions governing data release.

\subsection{Available Real results}\label{available-real-results}

\begin{table}[!htbp]
\centering\small
\setlength{\tabcolsep}{4pt}
\renewcommand{\arraystretch}{1.15}
\begin{tabularx}{\linewidth}{@{}Xrr@{}}
\toprule
Model & Correct / matched records & Accuracy \\

\midrule

GPT-4o & 54 / 237 & 22.78\% \\
Gemini 2.5 Flash & 62 / 237 & 26.16\% \\
Qwen2.5-VL-7B & 5 / 237 & 2.11\% \\
PIWM & 264 / 418 & 63.16\% \\

\bottomrule
\end{tabularx}
\caption{Available results matched to Real records.}
\end{table}

These are mapped, record-weighted results from
the existing real-store evaluation. The table reports each model's
available coverage separately.

\subsection{Reproduction record
structure}\label{reproduction-record-structure}

The experimental organization links task records, reference labels,
predictions, and scores by example identifier. A Full trace contains the
input observation identifier, predicted current state, each candidate
forecast, the final decision input, the raw decision output, and the
scored action. Ablation identifiers distinguish the four inference
configurations.

The following record map specifies the information needed to associate
each result with its evaluation unit.

\begin{table}[!htbp]
\centering\small
\setlength{\tabcolsep}{4pt}
\renewcommand{\arraystretch}{1.15}
\begin{tabularx}{\linewidth}{@{}lrX@{}}
\toprule
Evaluation & Records & Linked information \\

\midrule

See & 209 & Observation, stage and intent references, predicted state \\
Foresee & 309 & Current state, candidate action, next-stage reference
and prediction \\
Act & 245 & Textual reference state, candidate actions, chosen action \\
Full & 509 per configuration & Observation, state, forecasts, decision,
reference action \\
Real & 418 & Real-store observation, expert reference action,
decision \\

\bottomrule
\end{tabularx}
\caption{Record-level links for reproducing the current
evaluation.}
\end{table}

\subsection{Anonymous materials}\label{anonymous-materials}

Anonymous materials are hosted at:

\url{Code and data will be released after consent verification and upon publication.}

The hosted snapshot contains previously released data and tools. Its
historical results are separate from the current evaluation tables in
this paper. Snapshot updates require the current split manifests,
configurations, and prediction files.

Anonymous identifiers should remain consistent across inputs, labels,
and predictions. Media paths are relative to the release directory,
while private reverse mappings and credentials remain outside the
anonymous package.

\end{document}